\documentclass{article}
\usepackage{spconf,amsmath,epsfig}

\title{Knowledge-aware Visual Question Generation for Remote Sensing Images}
  \name{Siran Li, Li Mi\sthanks{corresponding author.}, Javiera Castillo-Navarro, Devis Tuia}
  \address{EPFL Switzerland}

\begin{document}
%\ninept
%
\maketitle
\begin{abstract}
With the rapid development of remote sensing image archives, asking questions about images has become an effective way of gathering specific information or performing image retrieval. However, automatically generated image-based questions tend to be simplistic and template-based, which hinders the real deployment of question answering or visual dialogue systems. To enrich and diversify the questions, we propose a knowledge-aware remote sensing visual question generation model, KRSVQG, that incorporates external knowledge related to the image content to improve the quality and contextual understanding of the generated questions. The model takes an image and a related knowledge triplet from external knowledge sources as inputs and leverages image captioning as an intermediary representation to enhance the image grounding of the generated questions. To assess the performance of KRSVQG, we utilized two datasets that we manually annotated: NWPU-300 and TextRS-300. Results on these two datasets demonstrate that KRSVQG outperforms existing methods and leads to knowledge-enriched questions, grounded in both image and domain knowledge.

\end{abstract}
\section{Introduction}
\label{sec:intro}

Extracting valuable and specific information from extensive archives of remote sensing images remains a challenge, especially for non-specialists. Asking questions in natural language can be an effective solution, enabling access to information of interest, while empowering the model to identify objects, gather spatial information, and interpret scenes~\cite{mostafazadeh2016generating, li2018visual, krishna2019information}. 
Visual Question Generation (VQG) for Remote Sensing Images, or RSVQG~\cite{bashmal2023visual}, aims at generating questions for remote sensing images, which plays a pivotal role in actively interacting with remote sensing data and provides possibilities to build a powerful Visual Question Answering (VQA) system or Visual Dialog system~\cite{lobry2020rsvqa}.

\begin{figure}[t]
    \centering
    \includegraphics[width=8.cm]{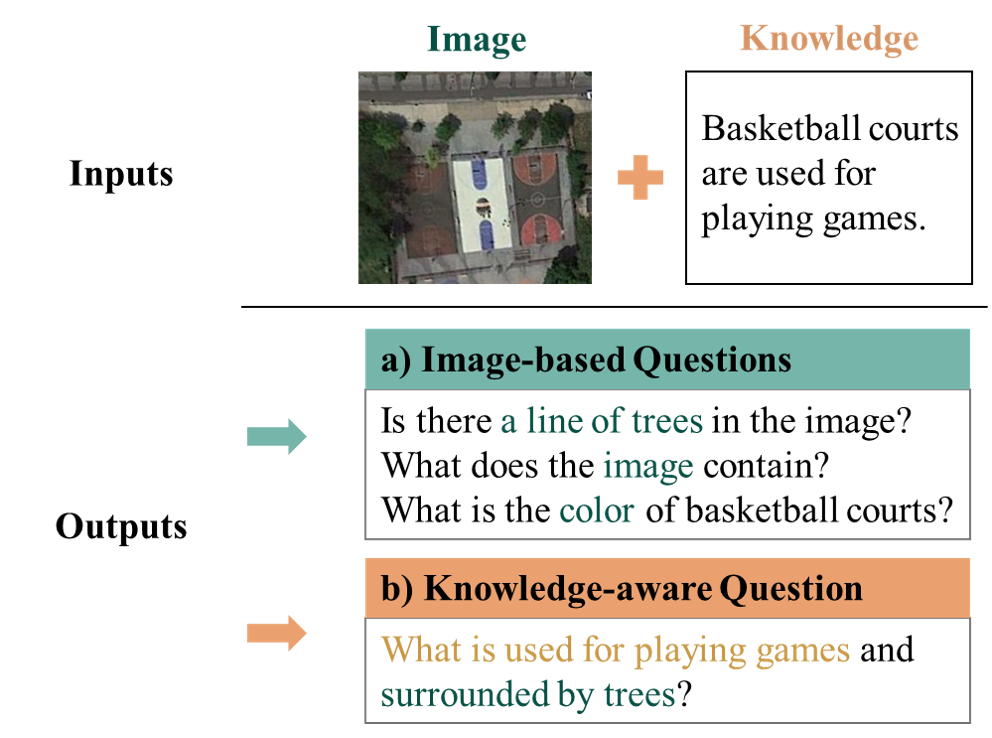}
    \caption{An example of \textbf{image-based questions} and a \textbf{knowledge-aware question}. In the questions, text from the image description is highlighted in green and text from external knowledge in orange.}
    \label{fig:sample}
\end{figure}
The quality and diversity of the questions generated are crucial to the success of such systems. Unfortunately, 
conventional image-based question generation systems tend to generate annotations that are redundant, template-based, and mostly focus on the presence of objects, rather than incorporating broader context or real-world information. Examples of such systems are shown in the green box of Fig.~\ref{fig:sample}, where generated questions are about objects (\textit{Is there a line of trees in the image?}) or very generic   (\textit{What does the image contain?}). These questions hardly meet the requirements for extracting information of interest and exploring commonsense beyond the raw image content (e.g. about the function of objects). To expand the questions beyond mere presence or counting ones, one might want to integrate external knowledge into VQG tasks, enabling the generation of questions that demand complex reasoning and yield informative insights \cite{gao2022cric}. This integration not only enhances the contextual understanding of the visual content, but also fosters specificity in questioning~\cite{xie2022knowledge, uehara2023k}. For example, knowledge-aware questions in the orange box of Fig.~\ref{fig:sample}   enhance specificity and relevance. They extract location information (\textit{surrounded by trees})  directly from the visual content and incorporate commonsense knowledge (\textit{Basketball courts are used for playing games.}) to construct more insightful and diverse questions.

% Current VQG models use an encoder-decoder structure to generate questions based on image content~\cite{li2018visual, patro2018multimodal}. To generate more specific questions, previous efforts proposed to add answers or question categories as additional inputs~\cite{krishna2019information, patil2020visual, xie2021multiple}. Furthermore, Uehara \textit{et. al} proposed to add knowledge triplets to generate knowledge-aware questions for natural images. While considerable attention has been devoted to VQG tasks on natural images, the RSVQG task is still in its infancy \cite{uehara2023k}. Bashmal \textit{et. al} introduced the VQG task for remote sensing images and used the remote sensing Visual Question Answering (VQA) datasets~\cite{lobry2020rsvqa, zheng2021mutual, bashmal2023visual} as the data source to generate image-based questions. However, as we discussed before, existing image-based questions in remote sensing predominantly feature questions eliciting simplistic yes/no or numeric responses, thus lacking the depth required for generating informative questions. 

\begin{figure*}[t!]
    \centering
    \includegraphics[width=17.5cm]{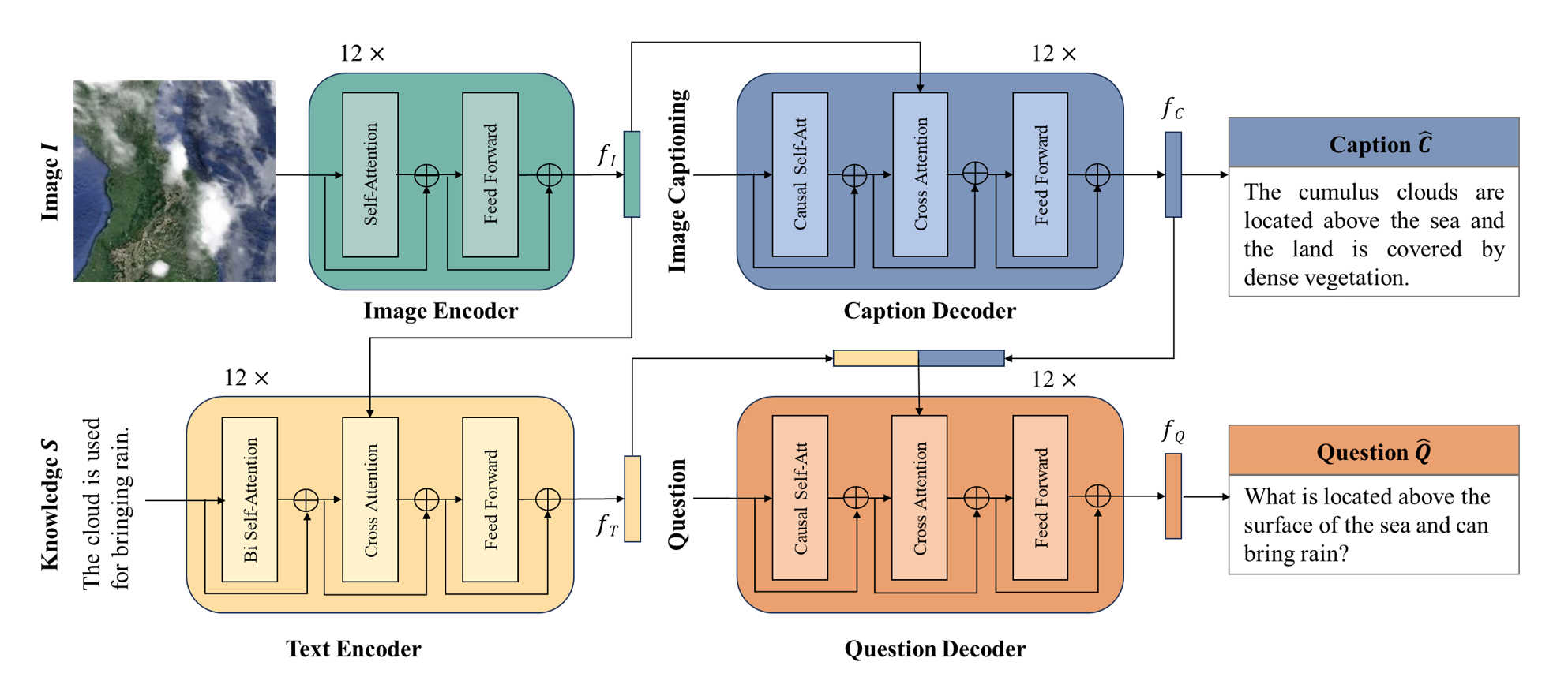}
    \caption{The pipeline of our KRSVQG model comprises four components: an image encoder, a caption decoder, a text encoder, and a question decoder. With an image ($I$) and a knowledge sentence ($S$) as inputs, the model generates a knowledge-aware question ($\widehat{Q}$) based on the caption ($\widehat{C}$) and the knowledge sentence ($S$). $\oplus$ means addition and normalization.}
    \label{fig:diagram}
\end{figure*}

%Moreover, as a dual task of VQA, the existing RSVQG task relies on remote sensing VQA datasets. The VQA datasets on remote sensing images~\cite{lobry2020rsvqa, zheng2021mutual} are considered answer-centric datasets, where questions are created using certain templates, resulting in fixed question patterns~\cite{chappuis2023curse}. %TextRS-VQA~\cite{bashmal2023visual}, while not restricted to fixed question types, still focuses on image-based questions with simple and generic content.  

To generate knowledge-enriched questions, we propose a Knowledge-aware RSVQG. The proposed KRSVQG method incorporates external knowledge to enrich the generated questions while leveraging descriptive captions as intermediary representations to enhance the image grounding of the generated questions. To evaluate KRSVQG, we use two datasets with 600 manually annotated samples in total (NWPU-300 and TextRS-300). The datasets are created by integrating external knowledge into questions, resulting in questions characterized by comprehensive and diverse content.

\section{Methodology}

Our KRSVQG model is built based on the BLIP structure introduced in Li \textit{et al.}~\cite{li2022blip} and consists of four components: an image encoder, a caption decoder, a text encoder, and a question decoder, as illustrated in Fig.~\ref{fig:diagram}. These components can be broadly classified into two modules. The first one is the vision module, which includes the image encoder and the caption decoder, responsible for encoding visual content. The second one is the language module, encompassing the text encoder and question decoder, for combining the knowledge input and image feature to generate questions. 

\subsection{Model Architecture}
% For the vision module, the BLIP image captioning model is used for extracting image feature representations.

\textbf{Image Encoder}.
Given an input image $I$, the image encoder based on Vision Transformer (ViT) \cite{dosovitskiy2020image} encodes the image features: $f_{I} = \mathbf{ViT}(I)$. 

{\noindent
\textbf{Caption Decoder}.
 The image feature is sent to the caption decoder to produce the caption feature: $f_{C} = \mathbf{CapDec}(f_{I})$. An explicit caption $\widehat{C}$ can be decoded based on the caption feature. The caption decoder incorporates a causal self-attention layer \cite{dong2019unified}, a cross-attention layer inserting visual feature $f_{C}$, and a feed forward network within each transformer block.
}

% \vspace{.2cm}\noindent As language module, we use a text encoder to process the input knowledge sentence and a  question decoder that fuses the image content with external knowledge to generate questions.

{\noindent
\textbf{Text Encoder}.
The text encoder shares a similar structure with the BLIP image-grounded text encoder~\cite{li2022blip}. Unlike the caption decoder's causal self-attention layers, the text encoder employs bidirectional self-attention layers to process the knowledge sentence $S$. The input knowledge sentence and the image feature $f_{I}$ are fused by the cross-attention layer and the resulting encoded feature is denoted as: $f_{T} = \mathbf{TextEnc}(S, f_{I})$.
}

{\noindent
\textbf{Question Decoder}.
The question decoder shares an identical structure with the caption decoder and injects the fused features of $f_{C}$ and $f_{T}$ by the cross-attention layer for generating the question: $f_{Q} = \mathbf{QueDec}(concat(f_{C}, f_{T}))$. The generated question sentence $\widehat{Q}$ can then be decoded from $f_{Q}$.
}

\subsection{Loss Functions}

To measure the dissimilarity between the predicted probability distribution and the target caption descriptions, we utilize a cross-entropy loss, denoted as caption generation loss:
\begin{equation}
\label{eq:cross-entropy}
\begin{aligned}
\mathbf{Loss}_{CG} = - \underset{n=1}{\sum^{|\widehat{C}|}}logP(\widehat{C}_{n}|\{\widehat{C}_{<n}\}),
\end{aligned}
\end{equation}
where $|\widehat{C}|$ is the number of tokens in the caption, $\widehat{C}_{n}$ is the token being generated at the n-th position, and $\widehat{C}_{<n}$ is the sequence of generated tokens up to the ($n-1$)-th step.

Similarly to caption generation, question generation also employs a cross-entropy loss to measure the dissimilarity between the predicted probability distributions and the target questions $Q$:
\begin{equation}
\label{eq:qgcross-entropy}
\begin{aligned}
\mathbf{Loss}_{QG} = - \underset{n=1}{\sum^{|\widehat{Q}|}}logP(\widehat{Q}_{n}|\{\widehat{Q}_{<n}\}),
\end{aligned}
\end{equation}
where $|\widehat{Q}|$ is the number of tokens in  $\widehat{Q}$, $\widehat{Q}_{n}$ is the token being generated at the n-th position, and $\widehat{Q}_{<n}$ is the sequence of generated tokens up to the ($n-1$)-th step.

The training process of our model follows three steps: (1) We pre-train the vision module using $\mathbf{Loss}_{CG}$ to adapt it to the remote sensing domain. (2) We pre-train the entire model on natural images (K-VQG dataset~\cite{uehara2023k}), to prepare the language module for knowledge-aware VQG. (3) We take the pre-trained vision module from (1), the pre-trained language module from (2), and fine-tune the entire model under $\mathbf{Loss}_{QG}$ supervision to generate knowledge-aware questions based on image captioning and the input knowledge sentence for remote sensing images.

\section{Datasets \label{datasets}}

To evaluate the proposed KRSVQG method, we build two knowledge-aware remote sensing VQG datasets: the NWPU-300 and the TextRS-300.

\textbf{Image Sources}. The datasets are based on samples from two existing remote sensing image captioning datasets: $300$ images from the NWPU dataset~\cite{cheng2022nwpu} and $300$ from the Text-RS dataset~\cite{abdullah2020textrs}, respectively. The NWPU caption dataset~\cite{cheng2022nwpu} is a large-scale dataset with diverse captions describing all prominent objects captured from various sensors, yielding resolutions between $30-0.2$m. While the TextRS dataset~\cite{abdullah2020textrs} offers captions with more general descriptions, primarily highlighting the predominant objects in the image. %This dataset collects images randomly selected from the AID, Merced, PatternNet, and NWPU datasets, providing a diverse range of visual data.

\textbf{Knowledge Source}. Our source for the knowledge sentence $S$ is ConceptNet~\cite{speer2017conceptnet}, a graph designed to represent commonsense knowledge and relationships between concepts. It contains over $1.7$ million entities and $3.4$ million edges as relationships. A typical triplet in ConceptNet is represented as \textit{$<$head, relationship, tail$>$}, such as \textit{$<$mobile houses, at location, street$>$}.

\begin{figure}[t!]
    \centering
    \includegraphics[width=8.5cm]{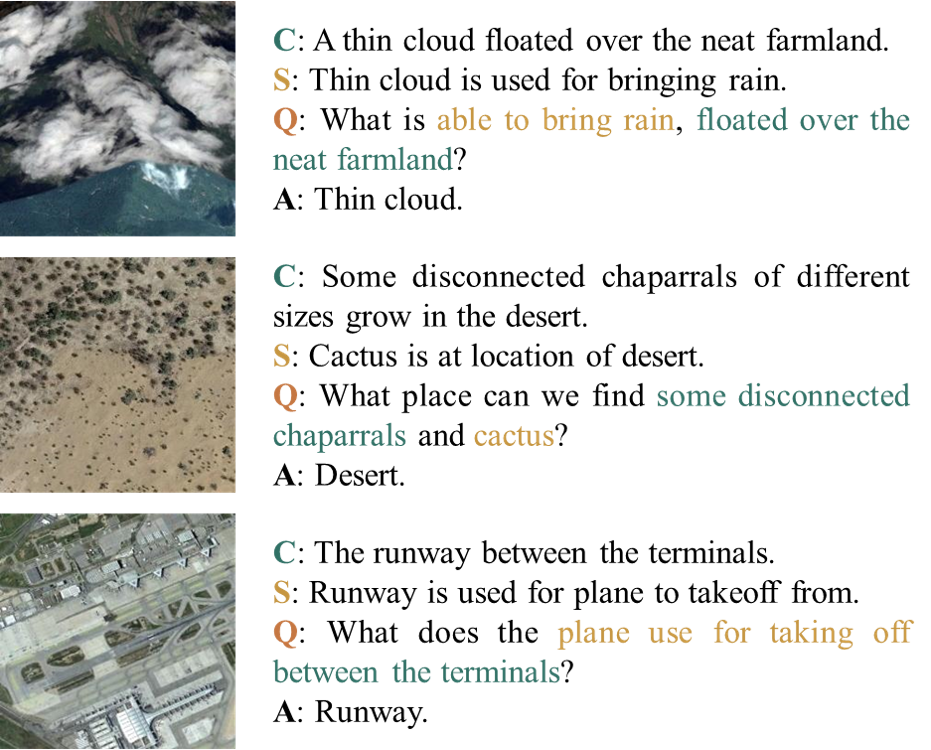}
    \caption{Example samples from the NWPU-300 and TextRS-300 datasets. 
    % We show the input images, caption (C), knowledge sentence (S), question (Q), and answer (A). 
    In the questions, text from the caption description is highlighted in green, and text from external knowledge is highlighted in orange.}
    \label{fig:data-ex}
\end{figure}

For constructing the datasets, we manually choose one triplet $T$ from ConceptNet. The triplet contains at least one object that is described in the image caption annotation to ensure that the selected triplet corresponds to the content of the image. Then we construct a concise sentence based on $T$, designated as knowledge sentence $S$; as answer $A$ to the question we sample one object appearing in the triplet (in the head or tail position). %Subsequently, we formulate a question $Q$, taking into account various elements, including $S$, $C$, $I$ and $A$. 
Following the above steps, the datasets consist of the following five components for each data sample: (1) image $I$, (2) caption $C$, (3) knowledge sentence $S$, (4) question $Q$, and (5) answer $A$, where images and captions are from the respective original remote sensing image captioning datasets. We partition the NWPU-300 and TextRS-300 datasets into training and validation sets, adhering to a $4:1$ ratio. Examples from both datasets are shown in Fig.~\ref{fig:data-ex}.

\section{Experimental Setup}

\subsection{Baseline Methods and Evaluation Metrics}

We take the following existing methods as the baselines:

\textbf{IM-VQG} \cite{krishna2019information}. The IM-VQG model is a VQG model that leverages variational auto-encoders to reconstruct both the image and answer representations,  helping the model to understand image content for question generation. For a fair comparison, we use knowledge $S$ as the answer input.
    
\textbf{AutoQG} \cite{ushio2022generative}. The AutoQG model is a sequence-to-sequence model designed to automatically generate questions from given paragraphs. In our experiments, we use the AutoQG model with T5-small architecture \cite{raffel2020exploring}. The text input is the concatenated knowledge sentence and caption.

Following the VQG evaluation in the literature~\cite{krishna2019information, bashmal2023visual}, we assess the performance of different question generation models using the following metrics: BLEU-$n$ ($n=1, 2, 3, 4)$ \cite{papineni2002bleu}, METEOR~\cite{banerjee2005meteor}, ROUGE$_{L}$~\cite{lin2004rouge}, and CIDEr~\cite{vedantam2015cider}. For all these metrics, a higher score indicates better performance.

% To facilitate model training and validation, we partition the NWPU-300 and TextRS-300 datasets into training and validation sets, maintaining a 4:1 ratio. Specifically, the training dataset comprises 240 samples, while the validation dataset encompasses 60 samples. It's worth noting that for the NWPU-300 validation dataset, annotations are derived from the NWPU validation dataset, which ensures that the VPT stage does not inadvertently reveal any information that could affect subsequent experiments.

\begin{table*}[t!]
    \centering
    \caption{\label{tab:finetune-rsdataset} Question generation results on the NWPU-300 and TextRS-300 datasets.}
    
    {\renewcommand{\arraystretch}{1.1}
    \begin{tabular}{c|cccccccc}
    \hline
    Dataset & Model & BLEU 1 & BLEU 2 & BLEU 3 & BLEU 4 & METEOR & ROUGE$_{L}$ & CIDEr \\\hline
    & IM-VQG~\cite{krishna2019information} & 32.46 & 15.91 & 7.93 & 0.00 & 8.96 & 31.95 & 0.17 \\ 
    NWPU- 300 & AutoQG~\cite{ushio2022generative} & 28.10 & 17.45 & 10.98 & 6.54 & 16.34 & 26.33 & 0.85 \\ 
    & \textbf{KRSVQG} & \textbf{41.87} & \textbf{28.32} & \textbf{20.73} & \textbf{14.78} & \textbf{18.70} & \textbf{38.48} & \textbf{1.24} \\\hline
    % & KRSVQG w/o CGD & 34.77 & 21.49 & 14.62 & 9.98 & 15.32 & 35.00 & 0.92 \\
    % & \textbf{KRSVQG} & \textbf{41.87} & \textbf{28.32} & \textbf{20.73} & \textbf{14.78} & \textbf{18.70} & \textbf{38.48} & \textbf{1.24} \\\hline
    & IM-VQG~\cite{krishna2019information} & 40.51 & 23.02 & 14.16 & 9.79 & 19.41 & 40.85 & 0.46 \\ 
    TextRS-300 & AutoQG~\cite{ushio2022generative} & 32.89 & 24.53 & 19.01 & 14.42 & \textbf{22.54} & 32.35 & \textbf{1.47} \\ 
    & \textbf{KRSVQG} & \textbf{44.26} & \textbf{33.78} & \textbf{27.63} & \textbf{22.90} & 19.64 & \textbf{42.90} & \textbf{1.47} \\ \hline
    \end{tabular}}
\end{table*}

\subsection{Implementation Details} 
We utilize a single NVIDIA GeForce RTX 2080 Ti GPU for our experiments. The BLIP base model with ViT-B and CapFilt-L, consists of 224M parameters, and the pre-trained weights are publicly available\footnote{https://github.com/salesforce/BLIP}. 
% For the following pre-training and fine-tuning process, the total training epoch of each step is set to 10, and the batch size is set to 1. The initial learning rate for the VPT and LPT is set to $2e^{-5}$, and for FT is set to $1e^{-6}$. 
We use the AdamW optimizer with a weight decay of 0.05, and the images are resized to $384 \times 384$ pixels.

\section{Results}
Question generation results on the NWPU-300 and the TextRS-300 dataset are reported in Table \ref{tab:finetune-rsdataset}. 

In both datasets, the IM-VQG~\cite{krishna2019information} model exhibits inferior performance when compared to both AutoQG and our KRSVQG model. This may be because external knowledge plays a crucial role in the task and IM-VQG is not designed for taking external knowledge as input. Even though the external knowledge is input for a fair comparison, its integration is not part of the model design, which we believe could affect the performance. 
    
The AutoQG~\cite{ushio2022generative} model outperforms the IM-VQG model, but it is still inferior to our proposed KRSVQG. As a language model, AutoQG does not use images as input and only uses captions and knowledge sentences. Its good performances show the importance of using knowledge information compared to a model not considering external information sources, but also shows its limitation due to the absence of visual information.
%demonstrates the importance of textual information in the generation of knowledge-aware questions.
    
Our KRSVQG model outperforms both the competing models on most metrics, with at least a relative $59$\% improvement on BLEU-4 on the NWPU-300 dataset and the TextRS-300 datasets and a $46$\% improvement on CIDEr for the NWPU-300 dataset. These improvements indicate that the KRSVQG model can capture the image content information and key concepts from the knowledge sentence more effectively than other models. 

% width=0.9\linewidth
\begin{figure}[t!]
    \centering
    \includegraphics[width=1\linewidth]{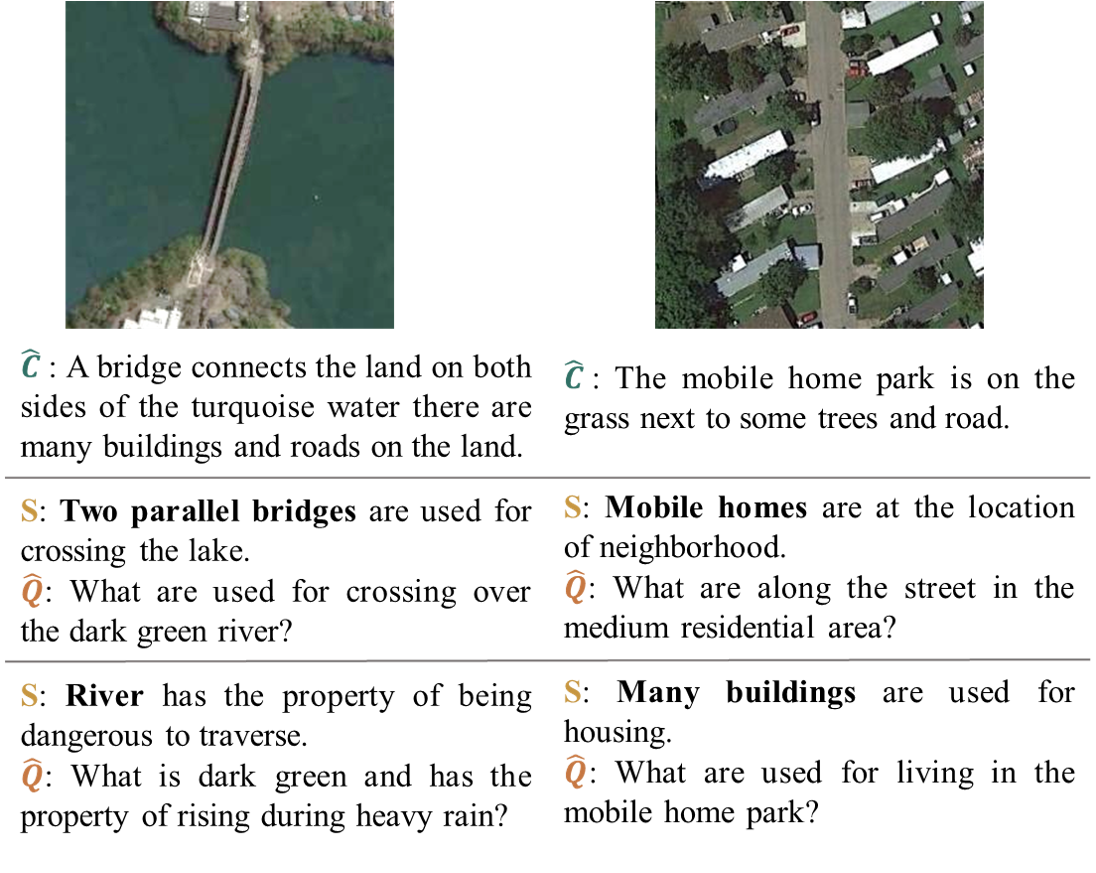}
    \caption{Two generated samples from the KRSVQG model on the NWPU-300 dataset. 
    % Each sample contains the input image, generated caption ($\widehat{C}$), two knowledge sentences ($S$), and the corresponding generated questions ($\widehat{Q}$). 
    The corresponding answer for $\widehat{Q}$ is marked in bold.
    }
    \label{fig:nwpu_qg}
\end{figure}

% \subsection{Qualitative Analysis}
Finally, in Fig. \ref{fig:nwpu_qg} we showcase a few captions and questions. KRSVQG generates them informed by knowledge, and the answer to each question is the object that is common between the caption and the input knowledge sentence. Such common object is highlighted in bold in $S$ in each example. Specifically, we provide two distinct knowledge sentences per image, along with their respective questions. The results highlight the model's ability to generate a wide range of questions based on varying knowledge inputs. For example, using the same image, different knowledge sentences such as \textit{Two parallel bridges are used for crossing the lake} and \textit{River has the property of being dangerous to traverse}, allow us to create questions from different perspectives.

\section{Conclusion}
In this paper, we propose a novel approach for generating knowledge-aware questions for remote sensing images, namely KRSVQG. The KRSVQG model incorporates knowledge sentences from external knowledge sources into the question generation process and uses image captioning as an intermediate stage for grounding the generated questions to specific image content. Results on two knowledge-aware remote sensing question generation datasets demonstrate the effectiveness of KRSVQG among the state-of-the-art VQG and QG methods in producing image-grounded and knowledge-enriched questions. In future research, we will use the generated questions in the VQA system to increase their generalization abilities and robustness.

\bibliographystyle{IEEEbib}
\bibliography{strings,refs}

\end{document}